\documentclass{article}

\usepackage{microtype}
\usepackage{graphicx}
\usepackage{subcaption}
\usepackage{booktabs} 

\usepackage{hyperref}

\usepackage[preprint] {icml2026}
\usepackage{multirow}

\usepackage{amsmath}
\usepackage{amssymb}
\usepackage{mathtools}
\usepackage{amsthm}
\usepackage{algorithm}
\usepackage{algorithmic}

\usepackage[capitalize,noabbrev]{cleveref}

\theoremstyle{plain}
\newtheorem{theorem}{Theorem}[section]

\theoremstyle{definition}

\theoremstyle{remark}
\newtheorem{remark}[theorem]{Remark}

\usepackage[textsize=tiny]{todonotes}

\icmltitlerunning{Topology-Aware Revival for Efficient Sparse Training}

\makeatletter
\icml@noticeprintedtrue

\renewcommand{\printAffiliationsAndNotice}[1]{}
\makeatother

\begin{document}

\raggedbottom

\twocolumn[
  \icmltitle{Topology-Aware Revival for Efficient Sparse Training
}

  \begin{center}
    {\large
      Meiling Jin\textsuperscript{1,*}\quad
      Fei Wang\textsuperscript{2,*}\quad
      Xiaoyun Yuan\textsuperscript{3}\quad
      Chen Qian\textsuperscript{1}\quad
      Yuan Cheng\textsuperscript{1,\#}
    \par}
    \vspace{0.35em}
    {\normalsize
      \textsuperscript{1} School of Artificial Intelligence, Shanghai Jiao Tong University, Shanghai, China\\
      \textsuperscript{2} SIMMIR Tech, Shanghai, China\\
      \textsuperscript{3} School of Computer Science, Shanghai Jiao Tong University, Shanghai, China\\
    }
    \vspace{0.35em}
    {\normalsize \texttt{jinmeilling@163.com}\par}
    \vspace{0.2em}
    {\footnotesize *Equal contribution\quad \#Corresponding author\par}
  \end{center}

  \icmlkeywords{network topology, static sparsity, pruning}

  \vskip 0.3in
]

\printAffiliationsAndNotice 

\begin{abstract}
Static sparse training is a promising route to efficient learning by committing to a fixed mask pattern, yet the constrained structure reduces robustness. 
Early pruning decisions can lock the network into a brittle structure that is difficult to escape, especially in deep reinforcement learning (RL) where the evolving policy continually shifts the training distribution.
We propose \textbf{Topology-Aware Revival (TAR)}, a lightweight one-shot post-pruning procedure that improves static sparsity without dynamic rewiring. After static pruning, TAR performs a single revival step by allocating a small reserve budget across layers according to topology needs, randomly uniformly  reactivating a few previously pruned connections within each layer, and then keeping the resulting connectivity fixed for the remainder of training.
Across multiple continuous-control tasks with SAC and TD3, TAR improves final return over static sparse baselines by up to \textbf{+37.9\%}  and also outperforms dynamic sparse training baselines with a median gain of \textbf{+13.5\%}.

\end{abstract}                                                                                                                                                                                                         
\begin{figure*}[t]
  \centering
  \includegraphics[width=\linewidth]{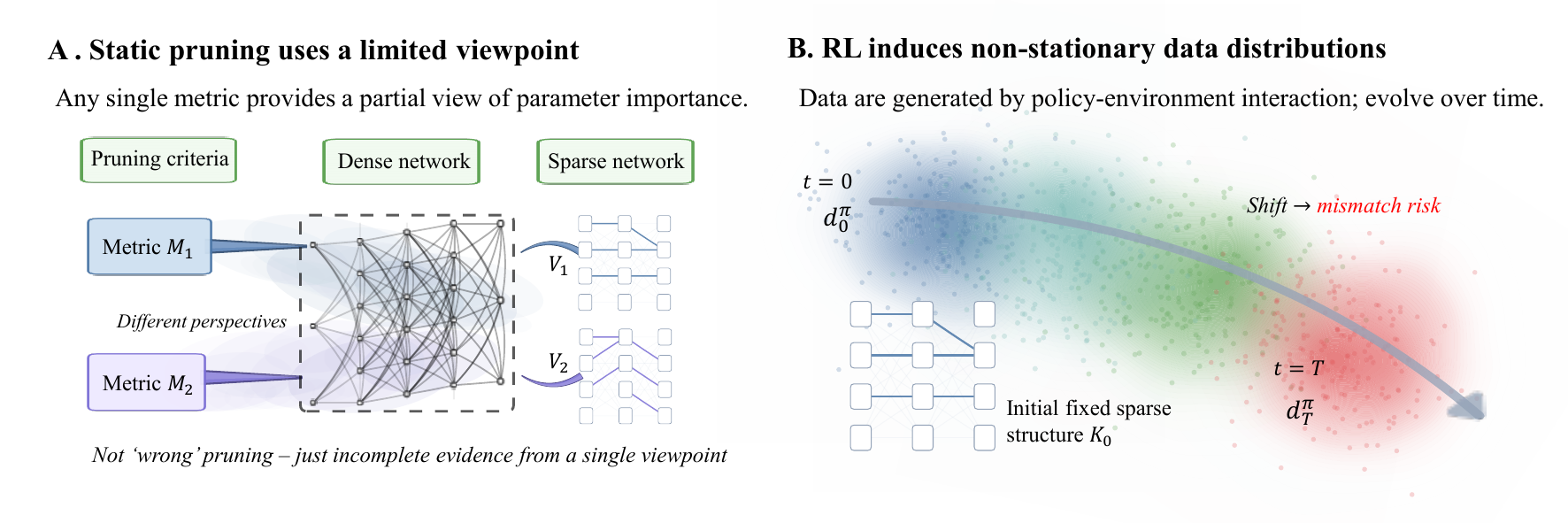}

  \caption{\textbf{Illustration of structural bottlenecks in static sparse training.} 
  \textbf{(A) Limited Viewpoint:} Any single pruning metric (e.g., $M_1$ or $M_2$) captures a particular viewpoint of importance, extracting a partial sparse structure ($V_1$ or $V_2$). Consequently, the resulting sparse structure represents an incomplete projection of parameter importance rather than a ``wrong'' one.
  \textbf{(B) Distribution shift in RL:} The policy-induced data distribution can drift from the initial stage ($d_0^{\pi_\theta}$) to the final stage ($d_T^{\pi_\theta}$). Initial fixed sparse structure $\mathcal{K}_0$, optimized solely for the early distribution, may fail to cover the requisite features for the shifted distribution, leading to mismatch risk.}
  
  \label{fig:static_pruning_viewpoint_shift}
\end{figure*}                                                                                                                                                                                                                                                                                                                            
\section{Introduction}
\label{sec:intro}

Static sparse training is a simple and effective way to sparsify neural networks: a sparsity mask is constructed once (often before training) and then kept fixed. It has delivered strong performance in supervised learning, including pruning at initialization~\citep{frankle2019lottery,WangQBZF22,0001GST24}, and is especially appealing in deep reinforcement learning (RL) as it avoids the overhead and complexity of dynamic sparse training. However, this same ``mask-once'' design also constrains the model structure throughout learning, reducing robustness to imperfect early pruning decisions. In RL, where the policy evolves and the data distribution drifts over training, a fixed sparse mask can be particularly brittle: at high sparsity, sparse agents may lag behind dense baselines or even collapse entirely \citep{graesser2022sparseRL}.

Figure~\ref{fig:static_pruning_viewpoint_shift} highlights two key reasons.
First, \textbf{static pruning criteria are limited by the network's available information} (Fig.~\ref{fig:static_pruning_viewpoint_shift}A). Early in training, the network has not yet learned all useful features. A pruning metric evaluates this partial state from a single perspective, yielding a fixed sparse structure that is reasonable but incomplete. For example, Magnitude keeps large weights \citep{han2015deep}, SynFlow preserves global signal propagation without data \citep{tanaka2020synflow}, and ERK-style allocation sets layer-wise sparsity from a structural prior \citep{dettmers2019sparse}. Each reflects a different viewpoint, and any single viewpoint can miss later-useful connections.

Second, \textbf{this brittleness is especially pronounced in RL} (Fig.~\ref{fig:static_pruning_viewpoint_shift}B). In RL, training data come from policy--environment interaction, so the data distribution drifts as the policy changes. The initially chosen mask is an \emph{early structural commitment}: it may fit the beginning distribution but miss connections needed later. When the distribution drifts, the fixed structure can become a persistent bottleneck because missing connections cannot be recovered.

\begin{figure*}[t]
  \centering
  \includegraphics[width=\linewidth]{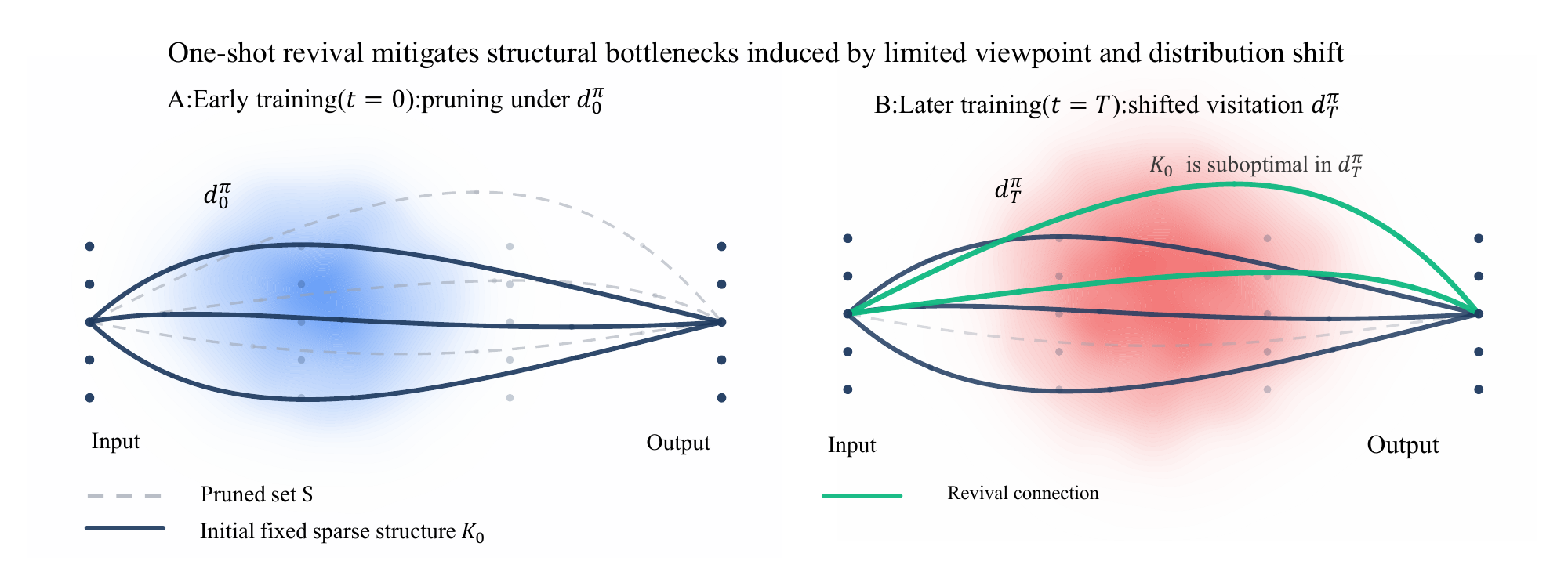}
\caption{\textbf{Revival mitigates distribution-shift-induced structural bottlenecks.}
  \textbf{(A) Early training ($t=0$):} static pruning selects a fixed sparse structure $\mathcal{K}_0$ under the early visitation distribution $d_0^{\pi_\theta}$, leaving pruned connections $\mathcal{D}$ (dashed).
  \textbf{(B) Later training ($t=T$):} as the visitation distribution drifts to $d_T^{\pi_\theta}$, the fixed sparse structure $\mathcal{K}_0$ may miss connections that become useful, creating a structural bottleneck.
  TAR performs a topology-aware one-shot revival that allocates a small reserve budget across layers based on connectivity needs and randomly revives a few pruned connections (green). The mask is then kept fixed, improving coverage and restoring gradient routes with minimal overhead.}
  \label{fig:revival_shift_bottleneck}
\end{figure*}

To address these problems, we propose \textbf{Topology-Aware Revival (TAR)}. Starting from a statically pruned network, TAR performs a \textit{one-shot revival}: it allocates a small ``revival budget'' across layers based on connectivity needs, then randomly uniformly  revives a few previously pruned connections in each layer. The revived edges act as \emph{reserve pathways}: they may stay inactive early, but they provide additional gradient pathways when the visitation distribution drifts (Fig.~\ref{fig:revival_shift_bottleneck}). Importantly, TAR keeps the connectivity fixed after this single step with minimal overhead.

Our contributions are summarized as follows:

(1) We analyze why static sparsity can be brittle and view static pruning as an \textbf{early structural commitment}. We highlight two drivers: (i) limited information and single-metric viewpoints, and (ii) policy-driven distribution drift, especially in RL.

(2) We introduce a \textbf{topology-aware one-shot revival} mechanism that allocates a small reserve budget across layers based on connectivity needs, revives a few pruned connections per layer, thereby improving layer-wise connectivity coverage under distribution drift.

(3) We evaluate across multiple representative continuous-control tasks with SAC and TD3, showing improved final return and stability: up to \textbf{+37.9\%} over static sparse baselines, and a median gain of \textbf{+13.5\%} over dynamic sparse training baselines, with negligible added cost.


\section{Related Work}
\label{sec:related}

\paragraph{Sparsity in neural network training.}
Research on network sparsity is well studied \citep{gale2019state,liu2019rethinking}. Most methods fall into two families: \emph{static sparse training} (SST) and \emph{dynamic sparse training} (DST) \citep{abs-1710-09282,BlalockOFG20}.

\textbf{SST} fixes a mask once (often before training), motivated by  lottery-ticket style ideas \citep{frankle2019lottery,zhou2019deconstructing,RendaFC20}. Common SST criteria include magnitude pruning \citep{han2015deep}, SNIP \citep{lee2019snip}, GraSP \citep{wang2020grasp}, and SynFlow \citep{tanaka2020synflow}. \textbf{DST} updates the mask during training by pruning and regrowing connection, with representative variants including SET and RigL \citep{mocanu2018set,evci2020rigl}, Deep Rewiring \citep{bellec2018deeprewire}, and dynamic sparse reparameterization \citep{mostafa2019dsr,GuoYC16}.

These methods are widely used as standard ways to sparsify neural networks.\citep{Sanh0R20} In comparison,SST is superior in its simplicity of implementation, while DST can update mask throughout training by pruning and regrowing edges.  This dynamic rewiring can reduce the risk of early structural commitment. However, dynamic rewiring comes with repeated mask updates (higher cost) and added complexity. 

This is especially challenging in RL: the data distribution already shifts as the policy changes, and DST introduces an additional source of non-stationarity by continually changing the network structure. Two-way dynamic changes may disrupt the continuity of network training and evolution.\citep{vincze2025smose}
We therefore ask a narrower question: \emph{can we keep the simplicity of static sparse training, but reduce the risk of early structural commitment under distribution drift?}

\paragraph{Non-stationarity and scaling in deep RL.}
RL training is non-stationary: the data distribution drifts as the policy evolves. This makes early pruning decisions harder to get optimal.

Furthermore, simply widening dense networks can yield limited gains and may introduce instability \citep{ota2021traininglarger,tang2025mitigating,wu2025dynamic}. In contrast, Ma et al.\ show that network sparsity can improve the scaling behavior of deep RL, and that appropriate sparse topologies can enable gains that do not reliably appear from widening dense networks \citep{ma2025network}.

Graesser et al.\ provide a systematic study of sparse training in deep RL and show that results can vary substantially across settings \citep{graesser2022sparseRL}.
Related analyses also link late-stage learning difficulties in deep RL to reduced plasticity \citep{lyle2023plasticity}, report dormant units and ways to restore capacity \citep{sokar2023dormant}, and highlight strong early-training effects (primacy bias) \citep{nikishin2022primacy}. These findings support our motivation: under non-stationarity, it helps to keep optional routes when the structure is fixed early.

\paragraph{Topology-aware sparsity.}
Topology-aware sparsity allocates sparsity using simple layer structure, rather than scoring individual weights.
ERK is a practical example: it uses fan-in/fan-out to set layer-wise sparsity \citep{dettmers2019sparse}.
More broadly, random graph theory gives basic intuition about connectivity thresholds \citep{bollobas2001randomgraphs}, and randomly wired structures can work well in practice \citep{xie2019randomly}. Motivated by these ideas, our TAR uses topology-aware budgeting, while keeping within-layer revival  randomly uniform and low cost.\citep{huang2025pruning}


\newcommand{\idxset}[1]{\{1,\dots,#1\}}

\section{Methodology}
\label{sec:method}

\subsection{Static Pruning as an Early Structural Commitment}
\label{sec:static_commitment }

\paragraph{Limited viewpoint in early pruning (Fig.~\ref{fig:static_pruning_viewpoint_shift}A).}

We consider static sparse training as a way to sparsify neural networks: a binary mask is constructed once and then kept fixed.
Let $d_t$ denote the training distribution at time $t$,  which characterizes the information feedback derived from the training data distribution. Static pruning selects a mask using early training signals (under $d_0$) and commits to that structure for the remainder of the training.

To induce sparsity, we apply a binary mask $\mathbf{m}\in\{0,1\}^d$ to the parameters $\theta$, where $m_i=1$ indicates that the parameter $i$ is kept and $m_i=0$ indicates that it is pruned.

The sparse parameters are
\begin{equation}
\label{eq:sparse_params}
\theta^{\text{sparse}}=\theta\odot \mathbf{m},
\end{equation}
where $\odot$ denotes element-wise multiplication.
We define network density $\rho$ and sparsity $\alpha$ as
\begin{equation}
\label{eq:density_sparsity}
\rho=\frac{\|\mathbf{m}\|_0}{d},
\qquad
\alpha=1-\rho,
\end{equation}
where $\|\mathbf{m}\|_0$ denotes $L_0$ norm of the binary mask $\mathbf{m}$, representing the count of non-zero(active) elements,and $d$ is the total number of parameters in the network.

Let $\mathcal{S}_t$ denote the optimal sub-network at time $t$,  which means a neural network structure that minimizes the loss under the current data set distribution; while  $^*$  denotes optimal  setting . That is, $\mathcal{S}^*_t$ stays constant if no changes to the dataset and tasks.
Let  $\mathcal{K}_t$  denote the reserved parameter set , with size $K_t=|\mathcal{K}_t|$; while let $\mathcal{D}_t$ be the set of pruned (dead) parameters, with size $D_t = |\mathcal{D}_t|$. Therefore, the initial fixed sparse network is made up of the parameters in $K_0$. For these parameters,
\begin{equation}
\label{eq:mask0}
m_t(i)=\mathbf{1}\{i\in \mathcal{K}_0\}.
\end{equation}
That is, initial fixed sparse structure $\mathcal{K}_0$  is 
\begin{equation}
\label{eq:static_training}
\theta^{\text{sparse}}_0=\theta\odot \mathbf{m}_t.
\end{equation}

Early in training, the network has not yet learned all useful features, while any metric evaluates the network from a single perspective: it scores parameters through one signal, and different signals emphasize different aspects.As the limited-viewpoint issue in Fig.~\ref{fig:static_pruning_viewpoint_shift}A shows, $\mathcal{K}_0$ and $\mathcal{S}^*_0$ can overlap when the pruning metric is sufficiently robust; however, in most cases, they usually remain distinct, existing marginal deviation.

Due to the fixed nature of the network structure, this marginal deviation cannot be ruled out, which  limits the ultimate performance of the network to some extent.

\paragraph{Mismatch especially in RL (Fig.~\ref{fig:static_pruning_viewpoint_shift}B).}

RL has a characteristic: training data come from policy--environment interaction, so the training distribution can change as the policy improves.

We consider a standard continuous control setting formulated as an MDP $\mathcal{M}=(\mathcal{S},\mathcal{A},P,r,\gamma)$ and a policy $\pi_\theta(a\mid s)$, where $\tau=(s_0,a_0,s_1,a_1,\dots)$ is a trajectory induced by $\pi_\theta$ in $\mathcal{M}$.
In RL,  a collection of multiple trajectories constitutes the dataset, which is obviously influenced by $\pi_\theta$ in $\mathcal{M}$, as the policy evolves.
Therefore,  in RL, the evolution of $d_t$ over time is driven not only by the model's feature extraction capabilities but also, to a greater extent, by the shifts in the dataset itself.

Let $d_t^{\pi_\theta}(s,a)$ denote the state--action visitation distribution at training time $t$.
As illustrated in Fig.~\ref{fig:static_pruning_viewpoint_shift}B, the early distribution $d_0^{\pi_\theta}$ and the later distribution $d_T^{\pi_\theta}$ generally exhibit significant differences; this distribution shift phenomenon is usually more pronounced when the task and the environment are complex.Thus, the optimal ideal sub-network $\mathcal{S}^*_t$ is not fixed but is changing. That is ,  $\mathcal{S}^*_0$ and  $\mathcal{S}^*_t$ do not completely overlap, indeed changing over time ; which means there is a larger deviation between $\mathcal{K}_0$ and $\mathcal{S}^*_t$ .

In conclusion, early in training the network's representation is still developing, so what the network can extract from data can change, which is universally applicable in neural network training. 
Second, in RL,  the visitation distribution $d_T^{\pi_\theta}$  can drift as the policy improves, shifting the data stream itself.
Putting the two together, from a model perspective, the effective training distribution $d_t$ changes significantly with training progress, in RL.

This leaves a fixed sparse structure vulnerable to mismatch.With a fixed mask, this mismatch persists and can become a structural bottleneck.

\begin{remark}[structure mismatch risk]
\label{rem:struct_under_coverage}
Static pruning is essentially a "premature" commitment to model structure made during the information-scarce early stages of training. In general deep learning, this risk stems from the inherent bias of initial evaluation metrics; in Reinforcement Learning (RL), this risk is exacerbated by the dynamic drift of data distributions caused by policy evolution. A fixed mask locks the model into an obsolete structure optimized for early-stage data, failing to cover critical connections required for later tasks, thus creating an insurmountable structural bottleneck. 
\end{remark}

\subsection{Topology-Aware Revival (TAR)}
\label{sec:TAR}

To mitigate this mismatch while retaining the characteristics of static pruning, we propose \textbf{TAR (Topology-Aware Revival)}. 
TAR performs a one-shot revival: it allocates a small “revival budget” across layers based on connectivity needs, then randomly uniformly revives a few previously pruned connections in each layer.

TAR acts as an  efficient one-shot mechanism to combat the risk of mismatch structure, by adding a few buffer parameters:they may stay inactive early, but provide additional gradient pathways when $\mathcal{K}_0$ is  insufficient.

In implementation, TAR involves two key questions: (i) layer-wise allocation (how many pruned parameters to revive per layer) and (ii) within-layer selection (which specific pruned parameters to revive).

\subsubsection{Topology-Aware Reserve Allocation}

To determine the revival budget, we analyze the structural properties of each layer. We model the weight matrix $W_\ell \in \mathbb{R}^{n^{\mathrm{out}}_\ell \times n^{\mathrm{in}}_\ell}$ of layer $\ell$ as a bipartite graph $\mathcal{G}_\ell = (\mathcal{V}_\ell, \mathcal{E}_\ell)$. Here, the vertex set $\mathcal{V}_\ell$ consists of two disjoint sets: the $n^{\mathrm{in}}_\ell$ input units and the $n^{\mathrm{out}}_\ell$ output units. The edge set $\mathcal{E}_\ell$ corresponds to the active parameters in $W_\ell$.

\paragraph{Topology Proxy and Connectivity Floor.}
We define a coarse node-count proxy, $N_\ell$, representing the total number of interacting units in the layer:
\begin{equation}
\label{eq:node_count}
N_\ell = n^{\mathrm{in}}_\ell + n^{\mathrm{out}}_\ell.
\end{equation}
In this graph-theoretic view, aggressive pruning risks creating isolated units, which severs gradient flow. Motivated by classical results in random graph theory (e.g., Erd\H{o}s--R\'enyi models), global connectivity typically emerges when the number of edges scales with $\Theta(N \ln N)$ \citep{bollobas2001randomgraphs}. Based on this, we establish a \textit{topology-aware floor}, $E_{\ell}^{\mathrm{topo}}$, as a minimum safety threshold for connectivity:
\begin{equation}
\label{eq:conn_floor}
E_{\ell}^{\mathrm{topo}} = \frac{N_\ell \ln N_\ell}{2}.
\end{equation}
The factor $\frac{1}{2}$ aligns with the critical threshold for random graphs: when the edge probability is $p \approx \ln N / N$, the expected number of edges is roughly $(N \ln N)/2$. Under this scale, the average node degree is proportional to $\ln N_\ell$, a regime where the probability of isolated units vanishes.

We use $E_{\ell}^{\mathrm{topo}}$ not as a strict guarantee, but as a budget-efficient signal to flag layers that are \emph{extremely} sparse. Let $K_\ell = \|\mathbf{m}_\ell^{(0)}\|_0$ denote the number of surviving edges (parameters) after the initial static pruning. If $K_\ell \ll E_{\ell}^{\mathrm{topo}}$, the layer is likely to contain poorly connected units and fragile gradient pathways. We quantify this risk via the \textbf{topology gap} $G_\ell$:
\begin{equation}
\label{eq:conn_gap}
G_\ell = \max\left(E_{\ell}^{\mathrm{topo}} - K_\ell, \ 0\right).
\end{equation}

\paragraph{Additional Protection and Final Allocation.}
To ensure structural plasticity across the entire network, we also employ a proportional recovery strategy. Let $\mathcal{D}_\ell$ be the set of pruned (dead) parameters, with size $D_\ell = |\mathcal{D}_\ell|$. We define a lightweight quota $Q_\ell$ based on a small recovery ratio($rr$):
\begin{equation}
\label{eq:quota}
Q_\ell = \lfloor rr \cdot D_\ell \rfloor.
\end{equation}
This ensures that even topologically robust layers receive a small exploration budget. TAR determines the final number of revived connections, $R_\ell$, by combining these two objectives:
\begin{equation}
\label{eq:final_rev_num}
R_\ell = \min\left(D_\ell, \ \max(G_\ell, Q_\ell)\right).
\end{equation}
This allocation strategy prioritizes layers that appear topologically under-connected (via $G_\ell$), while simultaneously ensuring every layer receives a minimum reserve (via $Q_\ell$) when $rr > 0$. In our implementation, the topology gap $G_\ell$ is applied only to matrix-like parameters (e.g., linear weights), whereas vector parameters (e.g., biases, LayerNorm) use $G_\ell=0$.

\subsubsection{Randomly uniform  Reserve Sampling}
\label{sec:rr_uniform}

Once the revival budget $R_\ell$ is determined for each layer via Eq.~\eqref{eq:final_rev_num}, the subsequent challenge is selecting \textit{which} specific pruned connections to revive. 
TAR adopts a minimalist yet theoretically grounded strategy: randomly uniform  sampling\textbf{ }within the pruned set $\mathcal{D}_\ell$.
This choice is necessitated by the two fundamental limitations discussed in Sec.~\ref{sec:static_commitment }:

(1) \textbf{Limited Viewpoint:} Early-stage metrics are inherently biased. Designing a specific scoring criterion based on $d_0$ to predict utility under $d_T$ is unreliable, as current "unimportant" edges may be latent features waiting for later activation.

(2)  \textbf{Non-stationarity in RL:} The distribution shift from $d_0^{\pi_\theta}$  to $d_T^{\pi_\theta}$  is driven by complex policy-environment interactions, making the future utility of specific edges stochastic and hard to predict.

Consequently, revival does not aim to recover edges that are \emph{currently} distinct; rather, it introduces optional gradient pathways to broaden the search space for future policy improvements.

\paragraph{Coverage Guarantee (Failure Probability Bound).}
Although the selection is random, we show that the probability of missing \textit{all} potentially useful connections decreases exponentially with the revival budget.

Let $\mathcal{S}^*_{T,\ell} \subseteq [d_\ell]$ denote the (unknown) oracle optimal sub-network required by the final task distribution $d_T$. 
We define the set of \textit{missed opportunities}—connections that are crucial for the future but were pruned initially—as:
\begin{equation}
\label{eq:missed_opp_set}
\mathcal{W}_\ell = \mathcal{S}^*_{T,\ell} \cap \mathcal{D}_\ell, \qquad w_\ell = |\mathcal{W}_\ell|.
\end{equation}
Here, $w_\ell$ represents the count of "latent gold" parameters hiding in the dead set $\mathcal{D}_\ell$.

Let $\mathcal{R}_\ell \subseteq \mathcal{D}_\ell$ be the set of revived parameters selected via randomly uniform sampling without replacement, with $|\mathcal{R}_\ell| = R_\ell$. 
We define the random variable $X_\ell = |\mathcal{R}_\ell \cap \mathcal{W}_\ell|$ as the number of useful connections successfully recovered. 
The probability of a\textbf{ }The probability of zero-recovery  (i.e., reviving $R_\ell$ parameters but hitting zero useful ones, $X_\ell = 0$) follows a hypergeometric distribution.

The probability is the ratio of "failed" combinations to total combinations:
\begin{equation}
\label{eq:hyper_geo_prob}
\Pr(X_\ell=0) = \frac{\binom{D_\ell-w_\ell}{R_\ell}}{\binom{D_\ell}{R_\ell}}.
\end{equation}
The numerator $\binom{D_\ell-w_\ell}{R_\ell}$ represents the number of ways to choose $R_\ell$ connections entirely from the $D_\ell - w_\ell$ "useless" parameters, while the denominator $\binom{D_\ell}{R_\ell}$ represents all possible ways to choose $R_\ell$ connections from the dead set.

To derive an interpretable upper bound, we approximate sa-mpling without replacement (hypergeometric) using sampling with replacement (binomial). Since sampling without replacement is strictly more efficient at exhausting non-targets, the binomial distribution provides a looser (upper) bound. Using the inequality $1 - x \le e^{-x}$, we obtain:
\begin{align}
\label{eq:miss_prob_bound}
\begin{split}
\Pr(X_\ell=0) 
&= \prod_{j=0}^{R_\ell-1} \left( 1 - \frac{w_\ell}{D_\ell - j} \right) \\
&\le \left( 1 - \frac{w_\ell}{D_\ell} \right)^{R_\ell} \le \exp\!\left( - \frac{R_\ell \cdot w_\ell}{D_\ell} \right).
\end{split}
\end{align}

\paragraph{Theoretical Implication.}
Eq.~\eqref{eq:miss_prob_bound} demonstrates that random revival acts as a strong probabilistic guarantee. The failure probability $\Pr(X_\ell=0)$ decays \textit{exponentially} as the revival budget $R_\ell$ increases. 
This implies that even if the density of latent useful parameters ($w_\ell / D_\ell$) is small, a modest increase in the revival budget $R_\ell$ (ensured by our quota $Q_\ell$) renders the probability of completely missing the sub-network $\mathcal{W}_\ell$ negligible.

Consequently, the random selection of specific parameters for revival is justified.

\paragraph{Computational overhead.}
TAR incurs negligible computational overhead. The one-shot  revival is performed exactly  immediately following the initial static pruning phase. Since the revived topology remains fixed, the per-iteration training cost is identical to that of standard static sparse training. 

\begin{algorithm}[t]
\caption{TAR: topology-aware one-shot revival (one-shot mask update, then fixed connectivity)}
\label{alg:revival}
\small
\begin{algorithmic}[1]
\REQUIRE Static masks $\{\mathbf{m}^{(0)}_\ell\}_{\ell=1}^{L}$, recovery ratio $rr$
\REQUIRE Layer shapes $\{(n^{\mathrm{in}}_\ell,n^{\mathrm{out}}_\ell)\}_{\ell=1}^{L}$
\ENSURE Revived masks $\{\tilde{\mathbf{m}}^{(0)}_\ell\}_{\ell=1}^{L}$
\FOR{$\ell=1,\dots,L$}
  \STATE $\mathcal{D}_\ell \gets \{i: m^{(0)}_{\ell,i}=0\}$; \quad $D_\ell \gets |\mathcal{D}_\ell|$
  \STATE $Q_\ell \gets \lfloor rr\, D_\ell \rfloor$
  \STATE $K_\ell \gets \|\mathbf{m}^{(0)}_\ell\|_0$
  \STATE $N_\ell \gets n^{\mathrm{in}}_\ell + n^{\mathrm{out}}_\ell$
  \STATE $G_\ell \gets \max\!\left(\frac{N_\ell\ln N_\ell}{2}-K_\ell,\,0\right)$
  \STATE $R_\ell \gets \min\!\big(D_\ell,\,\max(G_\ell,Q_\ell)\big)$
  \STATE Sample $\mathcal{R}_\ell \subseteq \mathcal{D}_\ell$ uniformly (w/o replacement) with $|\mathcal{R}_\ell|=R_\ell$
  \STATE $\tilde{\mathbf{m}}^{(0)}_\ell \gets \mathbf{m}^{(0)}_\ell$
  \FORALL{$i\in \mathcal{R}_\ell$}
    \STATE  $\tilde m^{(0)}_{\ell,i}\gets 1$
  \ENDFOR
\ENDFOR
\STATE \textbf{return} $\{\tilde{\mathbf{m}}^{(0)}_\ell\}_{\ell=1}^{L}$
\end{algorithmic}
\end{algorithm}

\section{Experiments}
\label{sec:experiments}

We design our empirical evaluation to answer four core research questions regarding the efficacy and robustness of Topology-Aware Revival:

RQ1 (Efficacy)\textbf{:} Does TAR effectively mitigate the structural mismatch risk? We assess whether TAR outperforms standard static pruning and uniform revival (UR) baselines across diverse algorithms and pruning criteria.

 RQ2 (Scalability): Can TAR scale to larger architectures? We test whether the topological benefits persist or amplify when network capacity increases (width $256 \to 1024$).
 
RQ3 (Mechanism Verification): Do the gains stem from improved topology or simply more parameters? We compare TAR against density-matched static controls to isolate the contribution of the revival mechanism.

RQ4 (Sensitivity): Is TAR robust to hyperparameter variations? We analyze the sensitivity of performance to the sparsity level $\alpha$ and recovery ratio $rr$.

We evaluate TAR on standard continuous control benchmarks using TD3 \citep{FujimotoHM18} and SAC \citep{HaarnojaZAL18}. Our primary testbed is Humanoid-v4, selected for its high dimensionality and structural complexity, alongside HalfCheetah-v4 and BipedalWalker-v3 for generalization.
We use standard MLPs (width 256) for both actor and critic networks. To ensure our method is agnostic to the pruning signal, we generate initial masks $\mathbf{m}^{(0)}$ using three established criteria: Magnitude, SynFlow, and ERK.
Comprehensive training details, including hyperparameters and environment specifications, are provided in Appendix~\ref{app:exp_details}.

\paragraph{Baselines.}
Given an initial mask $\mathbf{m}^{(0)}$ and a target sparsity $\alpha$, we compare the following strategies:

Original (Static): Standard static sparse training with no revival ($rr=0$).

    UR (Uniform Revival): A topology-agnostic baseline. For each layer $\ell$, it revives $R_\ell = \lfloor rr \cdot D_\ell \rfloor$ connections sampled uniformly from the pruned set $\mathcal{D}_\ell$.
    
    TAR (Ours): The proposed topology-aware revival, which allocates the budget based on structural gaps $G_\ell$ and proportional quotas $Q_\ell$.
    
    DST : Where feasible, we also report dynamic sparse training baselines (SET/RigL) for reference.


\begin{figure}[t]
  \centering
  \includegraphics[width=\linewidth]{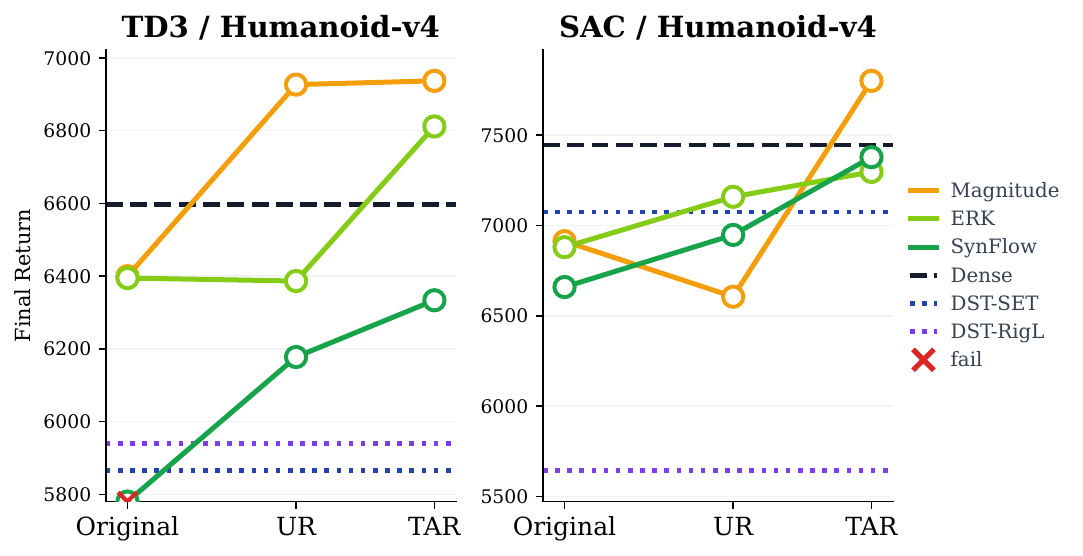}
\caption{\textbf{Humanoid-v4 (W=256): learning curves for TAR vs.\ Original (TD3/SAC) across static criteria.}
Each colored curve corresponds to a static criterion (Magnitude/ERK/SynFlow). Dashed lines are dense references; dotted lines are DST (SET/RigL) baselines.}
  \label{fig:rq1_humanoid_main}
\end{figure}

\begin{table*}[t]
\centering
\caption{\textbf{RQ1 (W=256): TAR vs.\ Original static across environments and criteria.}
Final return is mean$\pm$std over seeds (rounded to one decimal place).}
\label{tab:rq1_td3_sac_core3env}
\setlength{\tabcolsep}{3.2pt}
\renewcommand{\arraystretch}{1.08}
\scriptsize
\begin{tabular}{llccc|ccc cc}
\toprule
\multirow{2}{*}{\textbf{Criterion}} &
\multirow{2}{*}{\textbf{Variant}} &
\multicolumn{3}{c|}{\textbf{TD3 final return}} &
\multicolumn{3}{c}{\textbf{SAC final return}} &
\multirow{2}{*}{\textbf{$\alpha$ (TD3/SAC)}} &
\multirow{2}{*}{\textbf{$rr$}} \\
\cmidrule(lr){3-5}\cmidrule(lr){6-8}
& &
\textbf{HalfCheetah} & \textbf{BipedalWalker} & \textbf{Humanoid} &
\textbf{HalfCheetah} & \textbf{BipedalWalker} & \textbf{Humanoid} &
& \\
\midrule
\textbf{Dense} & &
8921.8$\pm$1045.6 & 318.4 $\pm$2.8& 6596.9$\pm$269.0 &
10092.6$\pm$936.9 & 315.1$\pm$15.7 & 7444.6$\pm$548.7 &
& \\
\midrule
\multirow{2}{*}{Magnitude} &
Original &
6818.6$\pm$704.4 & 297.6 $\pm$30.7& 6399.7$\pm$142.7 &
6984.8$\pm$115.0 & 314.2$\pm$10.8 & 6912.7$\pm$222.8 &
0.70/0.80 & 0.01 \\
& \textbf{TAR (ours)} &
\textbf{7244.9$\pm$805.6} & \textbf{317.1 $\pm$3.55}& \textbf{6936.9$\pm$119.9} &
\textbf{7516.0$\pm$286.3} & \textbf{324.3$\pm$2.3} & \textbf{7800.0$\pm$529.2} &
0.70/0.80 & 0.01 \\
\midrule
\multirow{2}{*}{SynFlow} &
Original &
7304.2$\pm$323.0 & 309.7$\pm$14.08& 342.8$\pm$16.2&
7939.2$\pm$311.1 & 315.9$\pm$7.5 & 6659.0$\pm$383.6 &
0.70/0.80 & 0.01 \\
& \textbf{TAR (ours)} &
\textbf{10204.3$\pm$1295.9}& \textbf{323.6$\pm$0.6}& \textbf{6333.6$\pm$185.7} &
\textbf{9652.3$\pm$1192.9} & \textbf{318.9$\pm$1.5} & \textbf{7377.4$\pm$166.7} &
0.70/0.80 & 0.01 \\
\midrule
\multirow{2}{*}{ERK} &
Original &
7402.1$\pm$648.1 & 318.2 $\pm$2.2& 6394.7$\pm$281.8 &
8660.4$\pm$576.6 & 324.8$\pm$2.9 & 6879.7$\pm$436.8 &
0.70/0.80 & 0.01 \\
& \textbf{TAR (ours)} &
\textbf{8082.2$\pm$669.0} & 315.7 $\pm$3.9& \textbf{6811.6$\pm$258.3} &
8447.0$\pm$1075.6 & \textbf{326.1$\pm$3.1} & \textbf{7296.1$\pm$694.0} &
0.70/0.80 & 0.01 \\
\midrule
\textbf{DST} & SET &
5256.0$\pm$2784.0 & 307.5 $\pm$13.6& 5865.8$\pm$348.7 &
6867.9$\pm$145.4 & 316.4$\pm$3.4 & 7074.4$\pm$374.5 &
0.70/0.80 & 0.01 \\
& RigL &
6478.9$\pm$183.8 & 298.6 $\pm$17.2& 5940.6$\pm$164.2 &
6948.6$\pm$560.8 & 321.5$\pm$4.4 & 5644.3$\pm$677.2 &
0.70/0.80 & 0.01 \\
\bottomrule
\end{tabular}
\end{table*}

\subsection{RQ1: Efficacy and Structural Robustness}
\label{sec:rq1_results}

We first analyze the performance of TAR across diverse environments and pruning criteria to validate its efficacy in mitigating structural mismatch.

\paragraph{TAR vs. Static Pruning (Mitigating Structural Bottlenecks).}
TAR consistently outperforms the static baseline (Original) across almost tested configurations, with relative performance up to $+37.9\%$ in SynFlow (TD3) (Table~\ref{tab:rq1_td3_sac_core3env}). This consistent improvement confirms that introducing a small, topology-aware reserve effectively alleviates the limitations of early structural commitment in RL.

\paragraph{TAR vs. Uniform Revival (Validating Topology Awareness).}
To isolate the benefit of our allocation strategy from the mere addition of parameters, we compare TAR against UR, which uses the exact same revival budget but samples connections uniformly in each layer. While UR improves over the static baseline in 4/6 settings (indicating that unstructured stochasticity offers some benefit against distribution shift), TAR significantly outperforms UR in 5/6 settings, with relative gains up to $+18.1\%$ in Magnitude(SAC) (Fig.~\ref{fig:rq1_humanoid_main}).
This result is crucial: it proves that the performance gains stem not just from "reviving parameters," but from \emph{where} they are revived. Allocating Revival budget is significant.

\paragraph{Comparison with Dynamic Sparse Training.}
Despite performing only a one-shot revival, TAR achieves competitive or superior performance compared to continuous dynamic sparse training methods (Table~\ref{tab:rq1_td3_sac_core3env}). This suggests that for RL tasks, a well-initialized static mask supplemented by a one-shot topology correction is a highly efficient alternative to expensive iterative topology updates.

\subsection{RQ2: Scalability to Larger Networks}
\label{sec:rq2_results}

\begin{figure}[t]
  \centering
  \includegraphics[width=\linewidth]{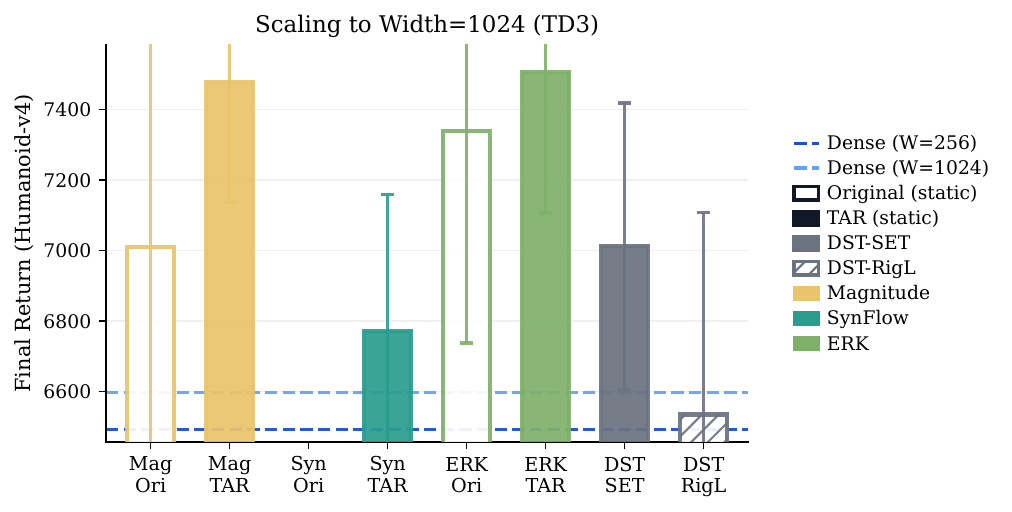}
  \caption{\textbf{RQ2 (TD3, Humanoid-v4): scaling to width 1024.}
  Blue dashed lines are dense references (W=256/1024). At W=1024 and $\alpha=0.8$, markers compare Original (circle) vs.\ TAR (square) for each static criterion; DST (SET/RigL) is shown for context (diamond).}
  \label{fig:rq2_td3_scaling}
\end{figure}

We investigate whether the topological benefits of TAR persist or amplify when the network capacity is quadrupled. We scale the MLP width from 256 to 1024 and evaluate performance on Humanoid-v4 (TD3) at a fixed sparsity of $\alpha=0.8$.

\paragraph{Limitations of Dense Scaling and Static Pruning.}
Consistent with prior observations in RL scaling, simply widening the dense network yields marginal performance gains. While standard static pruning (Original) can outperform the dense baseline under Magnitude and ERK criteria due to reduced overfitting or noise, it exhibits severe fragility under SynFlow, resulting in a complete performance collapse ($-96.9\%$ relative to dense).
This highlights that while static sparsity can aid scaling, the risk of early structural commitment increases with network size (Fig.~\ref{fig:rq2_td3_scaling}).

\paragraph{TAR Stabilizes Width Scaling.}
TAR effectively mitigates these scalability issues. It not only achieves a relative improvement of $+2.6\% \sim +13.8\%$ over the dense width-1024 reference across all criteria but also successfully rescues the policy from structural collapse in the SynFlow setting (restoring return from $206.6$ to $6770.2$). This confirms that a one-shot topology correction allows static sparse networks to utilize increased capacity reliably, preventing layer-level bottlenecks that standard static pruning fails to avoid.

\paragraph{Competitiveness with DST at Scale.}
Furthermore, TAR remains highly competitive with dynamic sparse training baselines at this scale, performing within the range of $-3.5\%$ (vs. SET) to $+14.9\%$ (vs. RigL) across criteria. This indicates that the simple, one-shot revival mechanism suffices for scaling to larger architectures, avoiding the implementation complexity and overhead of continuous topology updates required by DST.

\begin{table}[t]
\centering
\caption{\textbf{RQ3 (Humanoid-v4, TD3): density-matched control.}
Final return is mean$\pm$std over seeds (rounded to one decimal place).}
\label{tab:rq3_density_dst_td3}
\setlength{\tabcolsep}{3.2pt}
\renewcommand{\arraystretch}{1.08}
\scriptsize
\begin{tabular}{llccc}
\toprule
\textbf{Criterion} & \textbf{Variant} & \textbf{$\alpha_{\text{init}}$} & \textbf{$rr$} & \textbf{TD3 Return} \\
\midrule
\multirow{3}{*}{Magnitude}
& Original                 & 0.70 & --& 6399.7$\pm$142.7 
\\
& \textbf{TAR (ours)}     & 0.70 & 0.01 & \textbf{6936.9$\pm$119.9} \\
& Original (density-matched) & 0.69 & --& 6439.3$\pm$238.6\\
\midrule
\multirow{3}{*}{SynFlow}
& Original                 & 0.70 & --& 342.8$\pm$16.2\\
& \textbf{TAR (ours)}     & 0.70 & 0.01 & 
\textbf{6333.6$\pm$185.7}\\
& Original (density-matched) & 0.69 & --& 355.3$\pm$65.9\\
\midrule
\multirow{3}{*}{ERK}
& Original                 & 0.70 & --& 6394.7$\pm$281.8\\
& \textbf{TAR (ours)}     & 0.70 & 0.01 & \textbf{6811.6$\pm$258.3} \\
& Original (density-matched) & 0.69 & --& 6283.7$\pm$388.3 \\
\bottomrule
\end{tabular}
\end{table}


\subsection{RQ3: Mechanism Verification (Density-matched Control)}
\label{sec:exp_rq3}

To investigate whether TAR's performance gains stem from the algorithmic benefits of revival or simply from the marginal increase in parameter count (since revival slightly reduces sparsity), we introduce a density-matched static baseline.

\paragraph{Setup.}
With an initial sparsity $\alpha_{\text{init}}=0.70$ and recovery ratio $rr=0.01$, TAR results in a final sparsity of approximately $\alpha_{\text{init}}(1-rr) \approx 0.69$. Therefore, we construct a "Density-Matched Original" baseline initialized directly at $\alpha=0.69$ to ensure a fair comparison of model capacity.

\paragraph{Revival Mechanism vs. Parameter Count.}
As shown in Table~\ref{tab:rq3_density_dst_td3}, simply increasing the density of the static baseline yields negligible differences. The performance of Original at $\alpha=0.69$ deviates from $\alpha=0.70$ by only $-1.7\% \sim +3.7\%$ .
In contrast, TAR significantly outperforms the density-matched control across all criteria. For robust criteria (Magnitude/ERK), TAR achieves gains of $+7.7\% \sim +8.4\%$; for SynFlow, it restores performance to $6333.6$ from structural collapse. This confirms that the improvements are primarily driven by the \emph{topology-aware revival mechanism} correcting structural bottlenecks, rather than the trivial addition of parameters.


\subsection{RQ4: Sensitivity and Robustness}
\label{sec:rq4_results}

We assess the sensitivity of TAR to the recovery ratio $rr$ across varying sparsity levels $\alpha$, ensuring that our method does not rely on narrow hyperparameter tuning.

\begin{figure}[!ht]
  \centering
  \includegraphics[width=\linewidth]{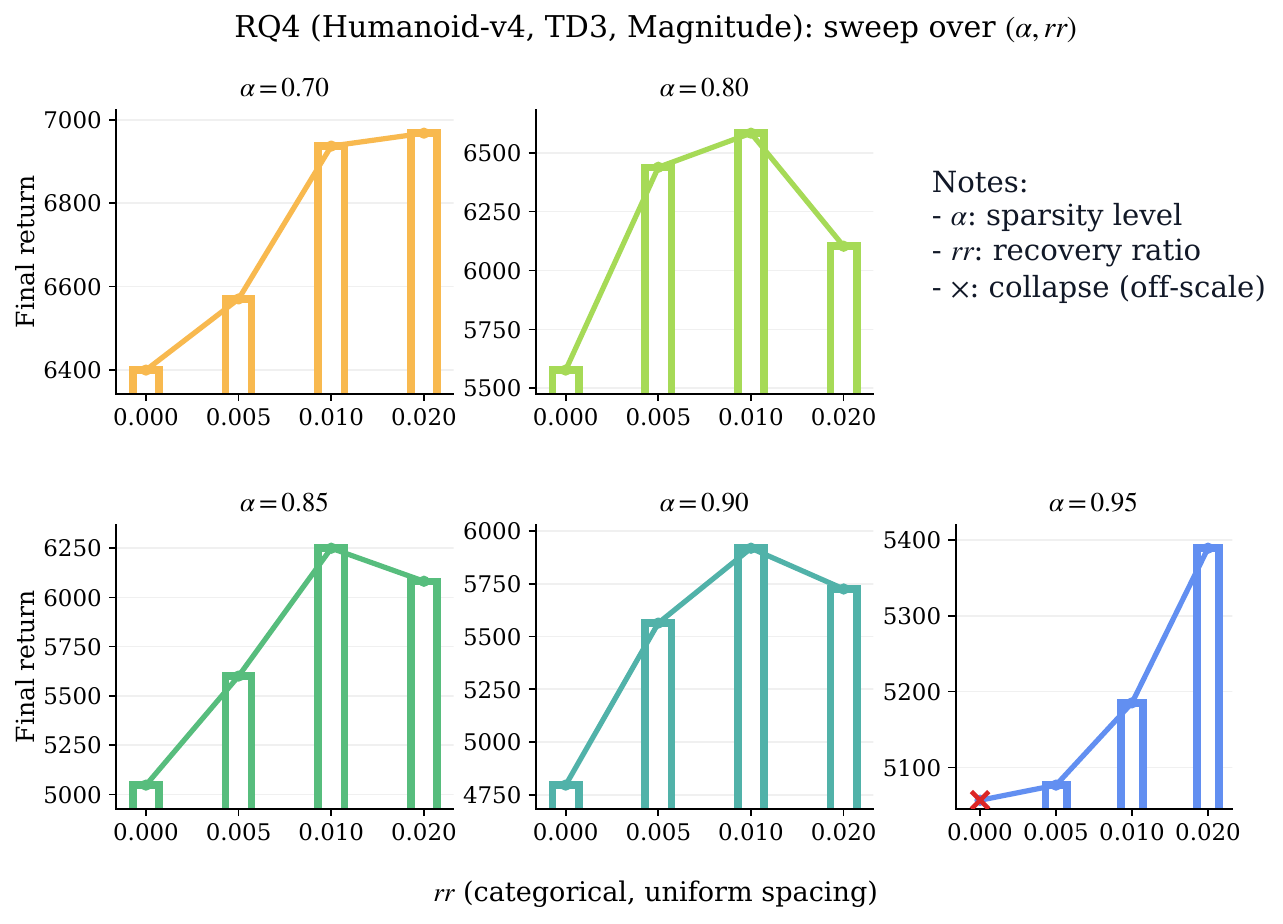}
  \caption{\textbf{RQ4 (Humanoid-v4, TD3, Magnitude):} sensitivity to recovery ratio $rr$ across sparsity levels $\alpha$. Each subplot fixes $\alpha$ and varies $rr\in\{0.000,0.005,0.010,0.020\}$ (categorical, uniformly spaced). Bars and overlaid lines report final return mean over seeds.}
  \label{fig:rq4_by_s_rr_bars_lines}
\end{figure}

\paragraph{Setup.}
We conduct a parameter sweep on Humanoid-v4 (TD3, Magnitude pruning) with sparsity levels $\alpha \in \{0.70, 0.80, 0.85, 0.90, 0.95\}$ and recovery ratios $rr \in \{0.0, 0.005, 0.01, 0.02\}$. Here, $rr=0$ represents the standard static pruning baseline.

\paragraph{Robustness across Sparsity Regimes.}
As shown in Fig.~\ref{fig:rq4_by_s_rr_bars_lines}, introducing a minimal revival budget ($rr > 0$) consistently improves performance over the static control ($rr=0$) across the entire sparsity spectrum. TAR exhibits strong robustness to the specific choice of $rr$; comparable gains are achieved across all non-zero values ($0.005 \sim 0.02$), indicating a stable "small reserve" regime that obviates the need for precise tuning. Notably, the benefits of TAR amplify at extreme sparsity levels ($\alpha \ge 0.90$). 
While the static backbone becomes increasingly brittle as redundancy decreases, the one-shot revival mechanism effectively compensates for structural deficiencies. Based on this stability, we adopt $rr=0.01$ as a conservative default for all main experiments, as it provides reliable protection against structural bottlenecks with negligible computational cost.

\section{Conclusion}
\label{sec:conclusion}
Static sparse training is a simple and efficient way to sparsify neural networks by fixing a mask initially, but this early structural commitment can reduce robustness.
This is especially problematic in RL, where the policy evolves and the training distribution drifts, so a fixed sparse structure can become a bottleneck later in training. 
We propose TAR, a lightweight one-shot post-pruning procedure that allocates a small reserve budget across layers, randomly uniformly revives a few pruned connections within each layer, and then keeps the mask fixed for the remainder of training.
Across TD3 and SAC on multiple continuous-control tasks, TAR improves final return over static sparse baselines by up to +37.9\% and outperforms dynamic sparse training baselines with a median gain of +13.5\%.
Ablations show that these gains are not explained by final density alone and persist when scaling network width. 
Overall, TAR preserves the main practical advantage of static sparse training: a fixed mask throughout training with zero mask-update overhead while improving robustness through a minimal intervention.
Future work may study how to design robust and efficient network topologies for non-stationary learning.

\section*{Impact Statement}
This paper presents work whose goal is to advance the field of Machine Learning. There are many potential societal consequences of our work, none which we feel must be specifically highlighted here.



\bibliography{example_paper}
\bibliographystyle{icml2026}

\newpage
\appendix
\onecolumn


\appendix

\section{Detailed Experimental Settings}
\label{app:exp_details}

\subsection{Environments and Training Protocols}
We conduct experiments on the MuJoCo continuous control suite (Gymnasium v26).Humanoid-v4: The primary high-dimensional environment. Trained for 5 million environment steps.HalfCheetah-v4 \& BipedalWalker-v3: Supplementary environments for robustness checks. Trained for 1 million environment steps.

We report the mean and standard deviation of the final return over 5 random seeds for Humanoid-v4 and 3 random seeds for the other tasks.

\subsection{Hyperparameters}
Table~\ref{tab:app_training_hparams_detailed} summarizes the specific hyperparameters used for TD3 and SAC algorithms. These settings are consistent with standard baselines in the literature to ensure fair comparison.

\begin{table}[h]
\centering
\caption{\textbf{Detailed hyperparameters for SAC and TD3.}}
\label{tab:app_training_hparams_detailed}
\setlength{\tabcolsep}{6pt}
\renewcommand{\arraystretch}{1.2}
\small
\begin{tabular}{lll}
\toprule
\textbf{Hyperparameter} & \textbf{SAC} & \textbf{TD3} \\
\midrule
Optimizer & Adam & Adam \\
Learning rate (Actor/Critic) & $3 \times 10^{-4}$ & $1 \times 10^{-4}$ \\
Hidden size & 256 & 256 \\
Batch size & 256 & 256 \\
Discount factor ($\gamma$) & 0.99 & 0.99 \\
Replay buffer size & $10^6$ & $10^6$ \\
Evaluation interval & 2500 steps & 2500 steps \\
Save interval & 100000 steps & 100000 steps \\
Update frequency & 50 env steps & 50 env steps \\
Start timesteps & 25000 & 25000 \\
\midrule
SAC entropy ($\alpha$) & 0.35 & -- \\
Target smoothing ($\tau$) & 0.005 & 0.005 \\
Exploration noise & -- & $\mathcal{N}(0, 0.15)$ \\
Policy delay freq & -- & 1 \\
\bottomrule
\end{tabular}
\end{table}

\begin{table*}[t]
\centering
\caption{\textbf{RQ1 (full matrix, W=256): multi-environment criterion scan.}
Final return is mean$\pm$std over seeds (rounded to one decimal place).
This table reports the full set (including partially completed runs), while the main paper focuses on the core \textbf{TAR vs.\ Original} comparison.}
\label{tab:rq1_full_matrix}
\setlength{\tabcolsep}{2.8pt}
\renewcommand{\arraystretch}{1.08}
\scriptsize
\begin{tabular}{llcccccccc}
\toprule
\multirow{2}{*}{\textbf{Criterion}} &
\multirow{2}{*}{\textbf{Variant}} &
\multicolumn{3}{c}{\textbf{TD3 final return}} &
\multicolumn{3}{c}{\textbf{SAC final return}} &
\multirow{2}{*}{\textbf{$\alpha$ (TD3/SAC)}} &
\multirow{2}{*}{\textbf{$rr$}} \\
\cmidrule(lr){3-5}\cmidrule(lr){6-8}
& &
\textbf{HalfCheetah} & \textbf{BipedalWalker} & \textbf{Humanoid} &
\textbf{HalfCheetah} & \textbf{BipedalWalker} & \textbf{Humanoid} &
& \\
\midrule
Dense & -- &
8921.8$\pm$1045.6 & 318.4$\pm$2.8 & 6596.9$\pm$269.0 &
10092.6$\pm$936.9 & 315.1$\pm$15.7 & 7444.6$\pm$548.7 &
-- & -- \\
\midrule
\multirow{3}{*}{Magnitude}
& Original & 6818.6$\pm$704.4 & 297.6$\pm$30.7 & 6399.7$\pm$142.7 &
6984.8$\pm$115.0 & 314.2$\pm$10.8 & 6912.7$\pm$222.8 &
0.70/0.80 & 0.01 \\
& +UR & 7033.6$\pm$370.3 &  310.68 $\pm$ 8.6& 6926.5$\pm$375.2 &
7240.6$\pm$155.4 & 313.1$\pm$7.6 & 6605.5$\pm$185.3 &
0.70/0.80 & 0.01 \\
& \textbf{+TAR} & \textbf{7244.9$\pm$805.6}& \textbf{317.1$\pm$3.6}& \textbf{6936.9$\pm$119.9} &
\textbf{7516.0$\pm$286.3}& \textbf{324.3$\pm$2.3} & \textbf{7800.0$\pm$529.2} &
0.70/0.80 & 0.01 \\
\midrule
\multirow{3}{*}{SynFlow}
& Original & 7304.2$\pm$323.0 & 309.7$\pm$14.1 & 342.8$\pm$16.2 &
7939.2$\pm$311.1 & 315.9$\pm$7.5 & 6659.0$\pm$383.6 &
0.70/0.80 & 0.01 \\
& +UR & \textbf{11206.0$\pm$354.0}& \textbf{326.7$\pm$2.6}& 6177.6$\pm$213.9 &
8619.8$\pm$699.2 & \textbf{319.9$\pm$4.1}& 6947.3$\pm$325.6 &
0.70/0.80 & 0.01 \\
& \textbf{+TAR} & 10204.3$\pm$1295.9& 323.6$\pm$0.6& \textbf{6333.6$\pm$185.7} &
\textbf{9652.3$\pm$1192.9} & 318.9$\pm$1.5 & \textbf{7377.4$\pm$166.7} &
0.70/0.80 & 0.01 \\
\midrule
\multirow{3}{*}{ERK}
& Original & 7402.1$\pm$648.1 & 318.2$\pm$2.2 & 6394.7$\pm$281.8 &
8660.4$\pm$576.6 & 324.8$\pm$2.9 & 6879.7$\pm$436.8 &
0.70/0.80 & 0.01 \\
& +UR & 7379.3$\pm$533.8 &  \textbf{320.0 $\pm$ 2.9}& 6386.5$\pm$527.4 &
\textbf{9149.4$\pm$80.9}& 322.4$\pm$1.5 & 7158.2$\pm$465.3 &
0.70/0.80 & 0.01 \\
& \textbf{+TAR} & \textbf{8082.2$\pm$669.0} & 315.7$\pm$3.9 & \textbf{6811.6$\pm$258.3} &
8447.0$\pm$1075.6 & \textbf{326.1$\pm$3.1} & \textbf{7296.1$\pm$694.0} &
0.70/0.80 & 0.01 \\
\midrule
DST (SET) & -- &
5256.0$\pm$2784.0 & 307.5$\pm$13.6 & 5865.8$\pm$348.7 &
6867.9$\pm$145.4 & 316.4$\pm$3.4 & 7074.4$\pm$374.5 &
0.70/0.80 & 0.01 \\
DST (RigL) & -- &
6478.9$\pm$183.8 & 298.6$\pm$17.2 & 5940.6$\pm$164.2 &
6948.6$\pm$560.8 & 321.5$\pm$4.4 & 5644.3$\pm$677.2 &
0.70/0.80 & 0.01 \\
\bottomrule
\end{tabular}
\end{table*}

\begin{figure}[t]
  \centering
  \includegraphics[width=0.98\linewidth]{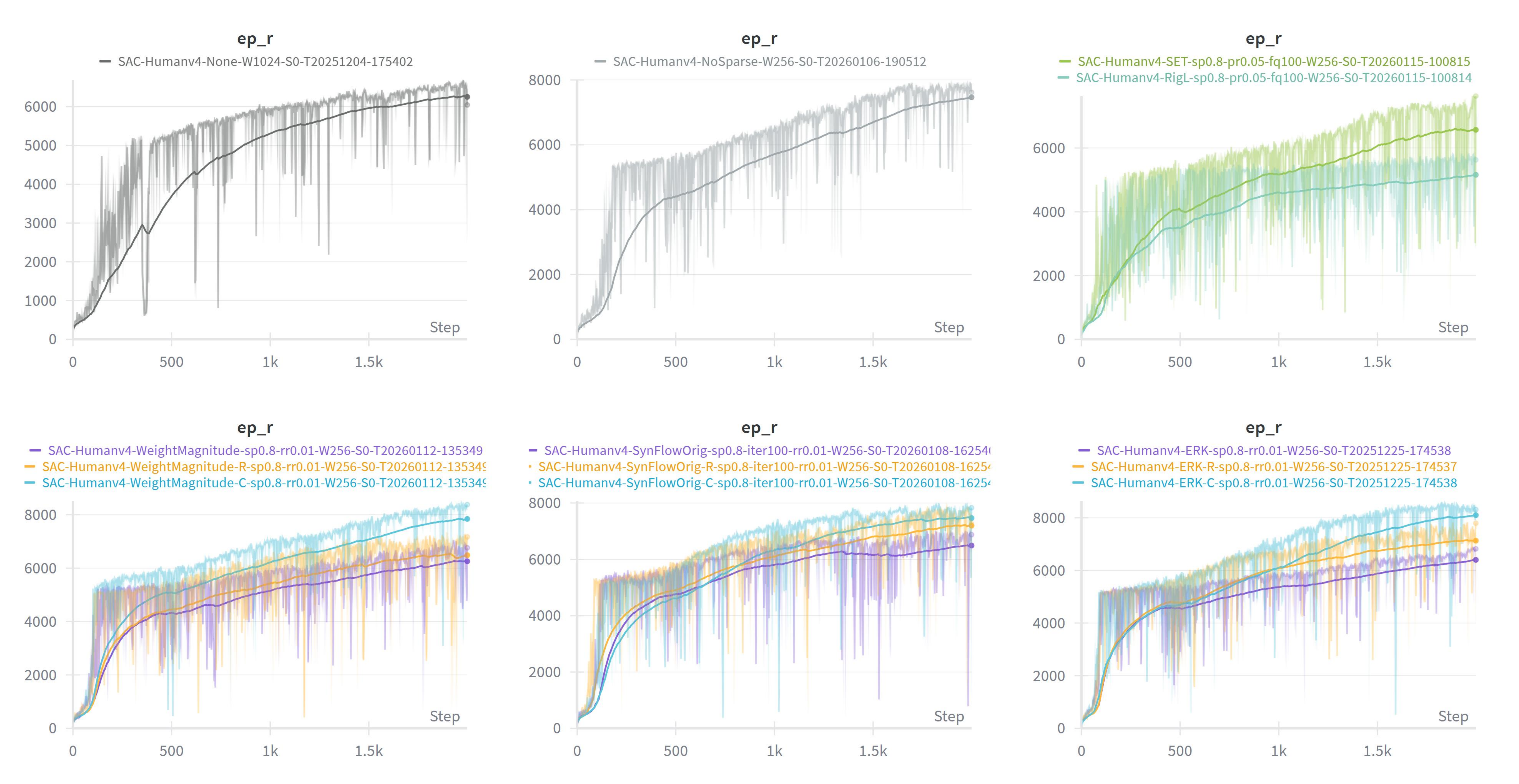}
  \caption{\textbf{Humanoid-v4 learning curves (SAC).}
  Same layout as Fig.~\ref{fig:app_wandb_grid_td3}, but for SAC.
  The grid highlights that UR can help but may be inconsistent, whereas TAR more reliably improves final return and stabilizes learning across criteria.}
  \label{fig:app_wandb_grid_sac}
\end{figure}

\section{UR (Uniform \emph{Random} Revival) as a same-budget ablation}
\label{app:rr_ablations}

\paragraph{Full RQ1 matrix including Original/UR/TAR.}
Table~\ref{tab:rq1_full_matrix} reports the complete RQ1 results (TD3/SAC, 3 environments, multiple criteria), including \textbf{Original}, \textbf{UR}, and \textbf{TAR}.
UR is the same-budget uniform revival control defined in Sec.~\ref{sec:experiments}.

\paragraph{Analysis and takeaway.}
The full matrix mirrors the main-text pattern: adding a small revival reserve (UR) often improves over Original and can prevent brittle failures, suggesting that a fixed early mask can be insufficient later in RL.
At the same time, TAR is typically stronger and more reliable than UR, indicating that allocating the same revival budget across layers matters in practice.
Together, Table~\ref{tab:rq1_full_matrix} and the main-text results show a consistent pattern: a single post-pruning reserve improves robustness under non-stationarity, and topology-aware budgeting often strengthens the effect.

\begin{figure}[t]
  \centering
  \includegraphics[width=0.98\linewidth]{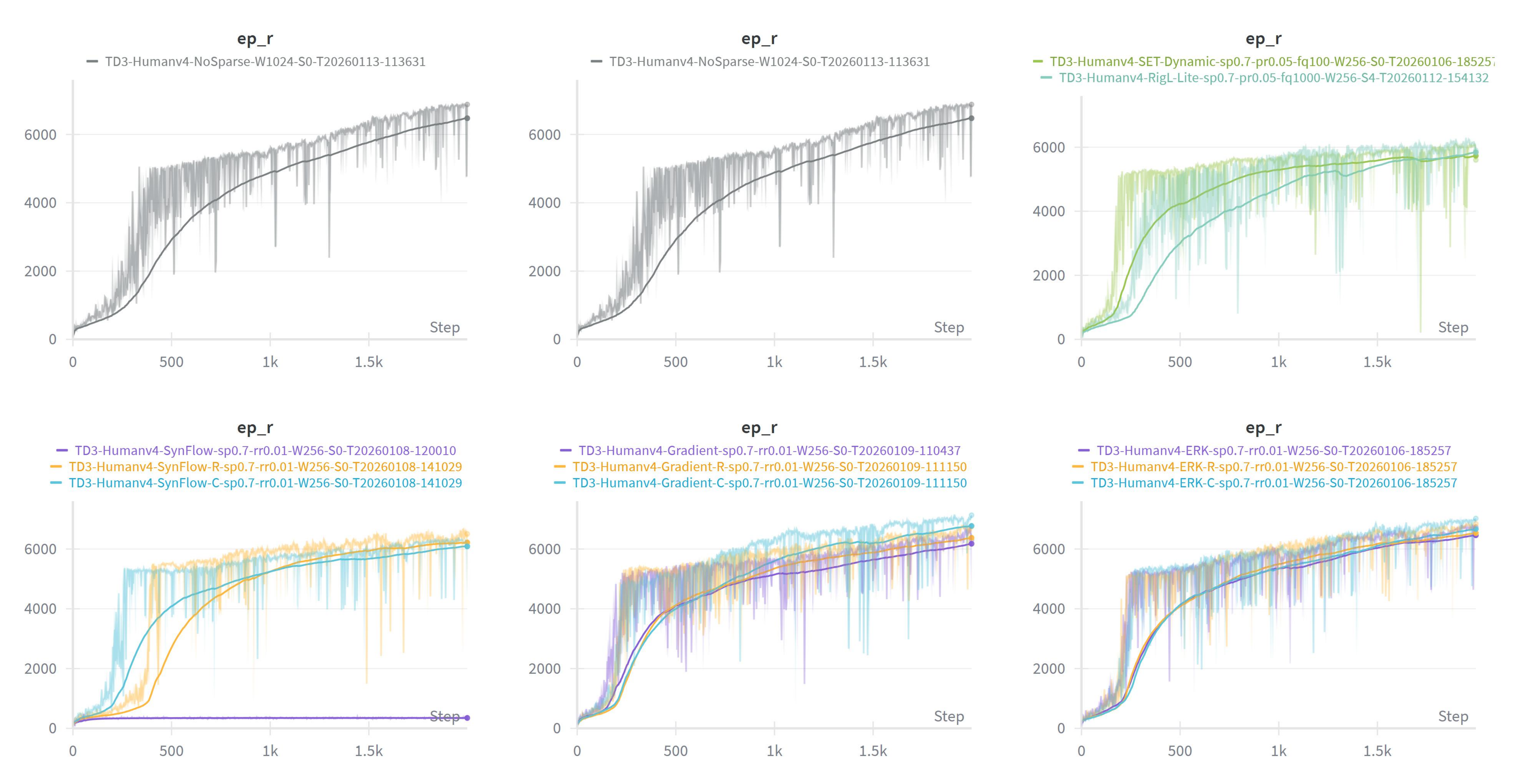}
  \caption{\textbf{Humanoid-v4 learning curves (TD3).}
  Episodic return vs.\ environment steps.
  Bottom row compares SynFlow / Gradient(Magnitude) / ERK under \textbf{Original} vs.\ \textbf{UR} vs.\ \textbf{TAR}; top row shows dense and DST baselines for context.
  UR can rescue brittle static backbones, while TAR is typically stronger/more stable.}
  \label{fig:app_wandb_grid_td3}
\end{figure}

\section{Learning curves (Humanoid-v4)}
\label{app:learning_curves}

Figures~\ref{fig:app_wandb_grid_td3}--\ref{fig:app_wandb_grid_sac} provide a compact learning-curve view on Humanoid-v4 for TD3 and SAC.
Each is a $2\times 3$ grid exported from W\&B: the top row gives dense/DST references, and the bottom row compares static backbones (Magnitude / SynFlow / ERK) under \textbf{Original} (static), \textbf{UR} (uniform random revival), and \textbf{TAR} (topology-aware revival).
Within each panel, the x-axis is environment steps and the y-axis is episodic return; shaded regions indicate uncertainty across seeds.
The bottom row is the key comparison: for each criterion, it shows whether a fixed early mask (Original) remains adequate throughout training, whether adding a uniform reserve (UR) prevents brittle failure, and whether TAR further improves stability and final return.

Across the grids, we see that static-sparse training can be unstable in some runs: an Original curve may stall far below other variants, while UR and TAR still learn normally (e.g., the bottom-left panel in Fig.~\ref{fig:app_wandb_grid_td3}).
When such brittleness appears, UR frequently improves the trajectory, consistent with a reserve buffer effect.
Across many panels, TAR tends to be the most stable among the three static-sparse variants and often ends with higher final return, suggesting that topology-aware budgeting helps the same reserve budget translate into more reliable learning.
Differences are often modest early but become clearer later in training, where distribution drift is stronger; this is consistent with reserve connectivity being most useful after visitation has shifted.
Overall, the learning curves align with the main-text takeaway: a single post-pruning reserve improves robustness under non-stationarity, and topology-aware allocation can further help.

\end{document}